# 眼球ロボットを用いた意図表出
# 及びマスコットロボットシステムへの応用
## Fuzzy Inference based Mentality Estimation for Eye Robot Agent

山崎 洋一　董 芳艶　増田 裕太　上原 由記子
Kormushev Petar　Vu Hai An　LE Phuc Quang　廣田 薫

Yoichi Yamazaki,　Fangyan Dong, Yuta Masuda, Yukiko Uehara,
Petar Kormushev, Hai An Vu, Phuc Quang Le, and Kaoru Hirota1

東京工業大学
Tokyo Institute of Technology

**Abstract:** Household robots need to communicate with human beings in a friendly fashion. To achieve better understanding of displayed information, an importance and a certainty of the information should be communicated together with the main information. The proposed intent expression system aims to convey this additional information using an eye robot. The eye motions are represented as states in a pleasure-arousal space model. Change of the model state is calculated by fuzzy inference according to the importance and certainty of the displayed information. This change influences the arousal-sleep coordinate in the space which corresponds to activeness in communication. The eye robot provides a basic interface for the mascot robot system which is an easy to understand information terminal for home environments in a humatronics society.

## １．はじめに

近年，ロボット技術が進歩するにつれてロボットが家庭環境へ普及することへの期待が高まっている．人間への親しみやすさが重要となる家庭環境下のロボットには，単にタスクをこなすだけでなく，親しみ易いインタフェースが必要である．またロボットは情報端末として，キーボードなどをインタフェースとする既存の情報端末よりも，人にやさしいインタフェースを持つものとなり得る．親しみやすさには心理表出が重要であり，心理表出には顔が重要であり，中でもとくに目が重要であることから，眼球ロボットによる心理表出の研究が行われている[1]．本研究では眼球ロボットを人間との対話インタフェースとして用いた室内分散型の家庭用情報端末ロボットアプリケーションであるマスコットロボットシステム，および提示する主情報に付随する情報を意図として表出する眼球ロボットによる意図表出を提案する．

## ２．マスコットロボットシステム

人間の生活空間において人間と共存するロボットは，エンターテインメントや高齢者支援などの観点で今後の発展が期待されている分野であるが，ロボットと人間の共存を考えた場合，発話などの明示的なコミュニケーションを行っていない時のロボットの振る舞いを考えることは重要である．

本研究では音声認識により目蓋と眼球を有する小型コミュニケーションロボットに対する人間の語りかけから人間の興味対象や嗜好を推定し，関連情報の提供を行うマスコットロボットシステム提案する．提案システムでは家庭環境において人間の興味対象や嗜好に関する情報を，室内に分散配置された小型コミュニケーションロボットが人間と生活を共にする中で獲得し，Web 情報推薦エンジンを利用して関連する情報を提供する．マスコットロボットシステムの機器構成を図１に示す．５台の小型コミュニケーションロボット，音声認識モジュール，情報推薦エンジンを RT ミドルウェアによって統合したマスコットロボットシステムを，家庭環境を想定した室内に構築する．小型コミュニケーションロボットとして５台の眼球ロボットを用いる．眼球ロボットを図２に示す．眼球ロボットは人間を参考にして２自由度の目蓋と３自由度の眼球を有する．音声認識モジュールの性能を補うため１台の眼球ロボットを自走型とする．自走型眼球ロボットを図３に示す．

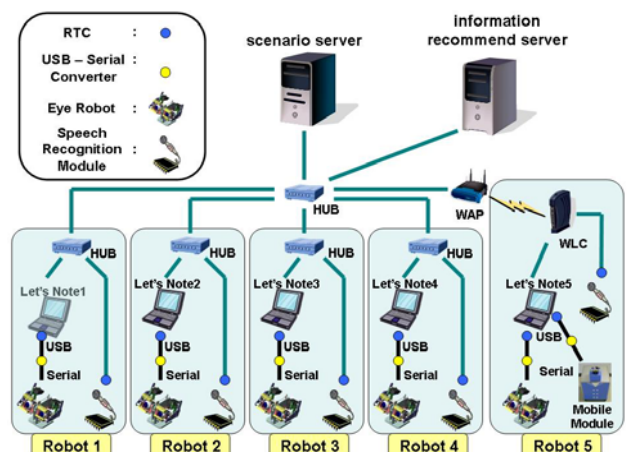

図１．マスコットロボットシステムの構成

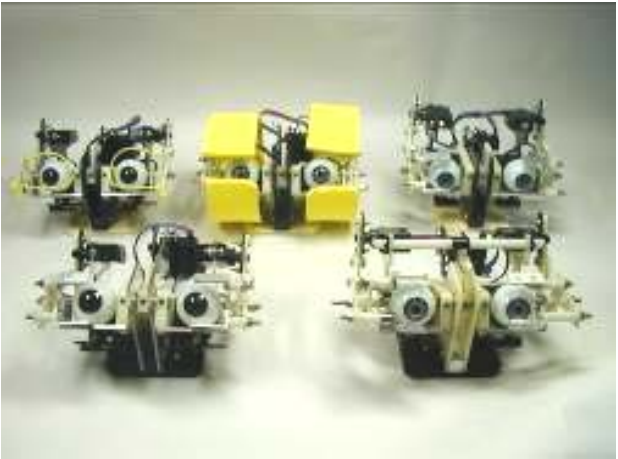

図2．5台の眼球ロボット

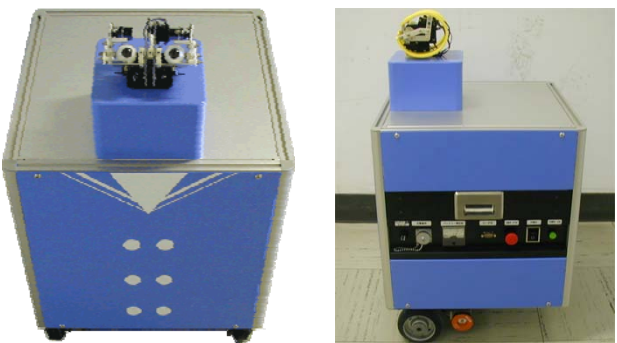

図3．自走型眼球ロボット

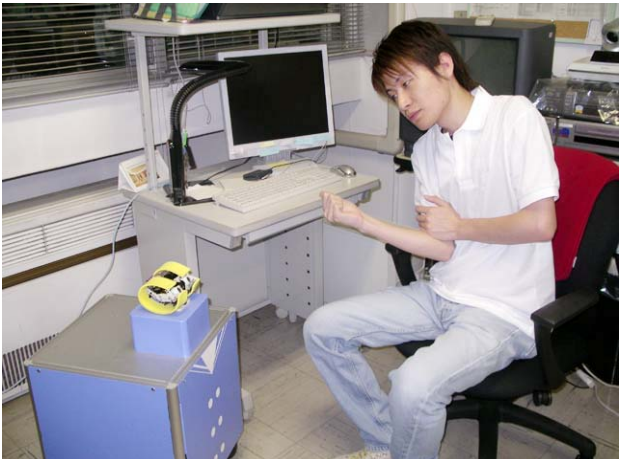

図4．ロボットとのコミュニケーション

## 3．意図表出

　人間とコミュニケーションするロボットアプリケーションを考える場合，認識性能の問題から認識に不確定な情報を含む可能性がある．また情報推薦に際して，優先順位付きの複数の結果を提示する可能性がある．このような状況では，主情報に付随する情報をロボットの意図として，主情報をともにさりげなく提示する必要がある．そこで提示情報とともにロボット自身の情報理解，提示情報の信頼性，重要性をわかりやすく提示するための眼球ロボットによる意図表出を提案する．

　意図表出と並ぶコミュニケーションロボットに必要な表出として，心理表出がすでに研究されている．眼による心理表出動作は単独では快－覚醒平面上に配置される．快－覚醒平面は対話への積極性を表す「覚醒－睡眠」軸と対話者への即時的な好意を表す「快－不快」軸の2軸からなる．コミュニケーションにおける心理状態の推移を考慮し，親和の1軸を追加した親和型快－覚醒空間が提案されている[1]．

　意図表出では音声認識により入力された情報に対して，入力に対する確信度，出力に対する信頼度，重要度を算出しあわせて，「覚醒－睡眠」軸に投影することにより心理表出を加味した意図表出を実現する．

## 4．おわりに

　家庭環境で人間と共存するロボットアプリケーションとしてマスコットロボットシステムを提案し，その意図表出システムを構築する．マスコットロボットシステムでは，5台の小型コミュニケーションロボット，音声認識モジュール，情報推薦エンジンをRTミドルウェアによって統合する．眼球ロボットは人間を参考にして2自由度の目蓋と3自由度の眼球を有し，5台のうち1台を自走型にする．人間とのコミュニケーションを考慮し，提示する主情報に付随する情報を意図として表出する眼球ロボットによる意図表出を提案する．入力された情報に対して，入力に対する確信度，出力に対する信頼度，重要度を算出しあわせて，「覚醒－睡眠」軸に投影することにより心理表出を加味した意図表出を実現する．

　提案システムでは，インタフェースに操作の習得を必要とするキーボードなどではなく，人間同士のコミュニケーションとおなじ音声認識を用いている．マスコットロボットシステムは既存の情報端末に比べて人間やさしいインタフェースを持つロボットシステムとして，家庭に普及し得る．

## 連絡先

山崎　洋一
E-mail: yama@hrt.dis.titech.ac.jp